\documentclass{bmvc2k}
\usepackage{csvsimple}
\usepackage{multirow}
\usepackage{array}
\usepackage{enumitem}
\usepackage{lipsum}
\usepackage{amsmath}
\newcolumntype{L}[1]{>{\raggedright\let\newline\\\arraybackslash\hspace{0pt}}m{#1}}
\newcolumntype{C}[1]{>{\centering\let\newline\\\arraybackslash\hspace{0pt}}m{#1}}
\newcolumntype{R}[1]{>{\raggedleft\let\newline\\\arraybackslash\hspace{0pt}}m{#1}}

\title{Features for Ground Texture Based Localization - A Survey}

\addauthor{Jan Fabian Schmid}{SchmidJanFabian@gmail.com}{1}
\addauthor{Stephan F. Simon}{}{1}
\addauthor{Rudolf Mester}{}{2}

\addinstitution{
	Robert Bosch GmbH\\
	Hildesheim, Germany
}
\addinstitution{
	Norwegian Open AI Lab\\
	Computer Science Dept. (IDI)\\
	NTNU Trondheim, Norway
}

\runninghead{Schmid, Simon, Mester}{Ground Texture Feature Survey}

\def\etal{\emph{et al}\bmvaOneDot}

\begin{document}

\maketitle

\begin{abstract}	
	Ground texture based vehicle localization using feature-based methods is a promising approach to achieve infrastructure-free high-accuracy localization.
	In this paper, we provide the first extensive evaluation of available feature extraction methods for this task,
	using separately taken image pairs as well as synthetic transformations.
	We identify AKAZE, SURF and CenSurE as best performing keypoint detectors, 
	and find pairings of CenSurE with the ORB, BRIEF and LATCH feature descriptors to achieve greatest success rates for incremental localization,
	while SIFT stands out when considering severe synthetic transformations as they might occur during absolute localization.
\end{abstract}

\section{Introduction}
\label{sec:introduction}
Highly accurate localization capabilities are required to enable the use of autonomous robots and vehicles.
Available solutions such as GPS for outdoor applications are not able
to reliably provide accurate positioning in urban environments,
and systems for indoor applications such as Ultra Wideband require installation of costly infrastructure
\cite{Cornick_Radar, chenstreetmap, Fang_intelligent-vehicles2}.
Visual localization using environmental landmarks can achieve centimeter precise localization in some indoor applications,
but might suffer from occlusions and can deviate meters from the correct position in outdoor scenarios~\cite{Mur-Artal_ORBSLAM2}.
Localization approaches based on ground texture using a downward-facing camera, on the other hand,
present promising results for reliable centimeter precise localization~\cite{Zhang_High-Prec-Localization, chenstreetmap}.
Suitable texture types like concrete, asphalt, or carpet are prevalent and remain sufficiently stable in most application areas of autonomous agents \cite{Zhang_High-Prec-Localization, Kelly_AGV}.
Therefore, ground texture based solutions have the potential to enable infrastructure-free high-accuracy localization.
Furthermore, they enable localization in environments without static landmarks and
can help to avoid privacy issues of household robots.

\begin{figure*}
	\centering
	\includegraphics[width=0.1614\columnwidth]{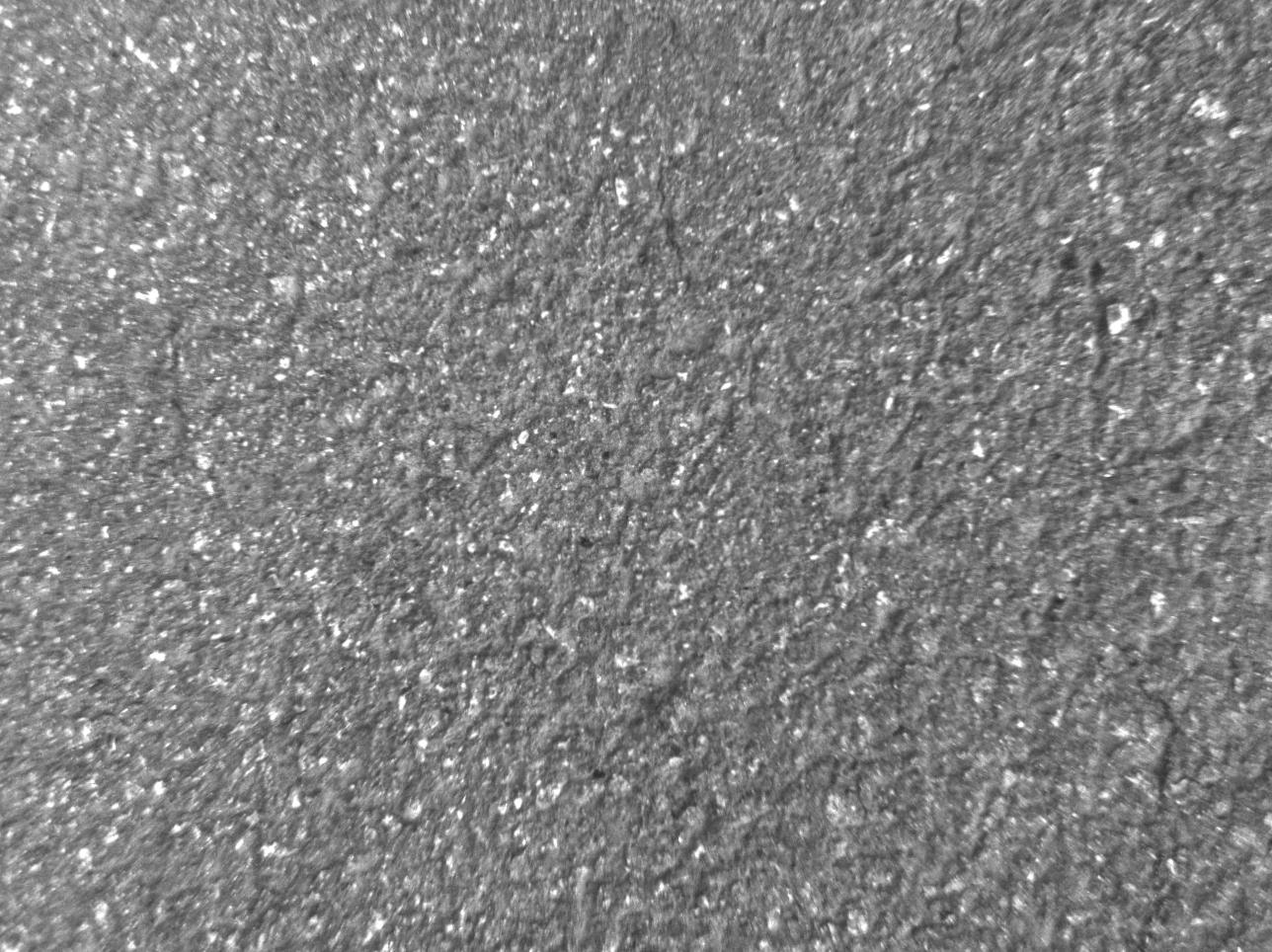}
	\includegraphics[width=0.1614\columnwidth]{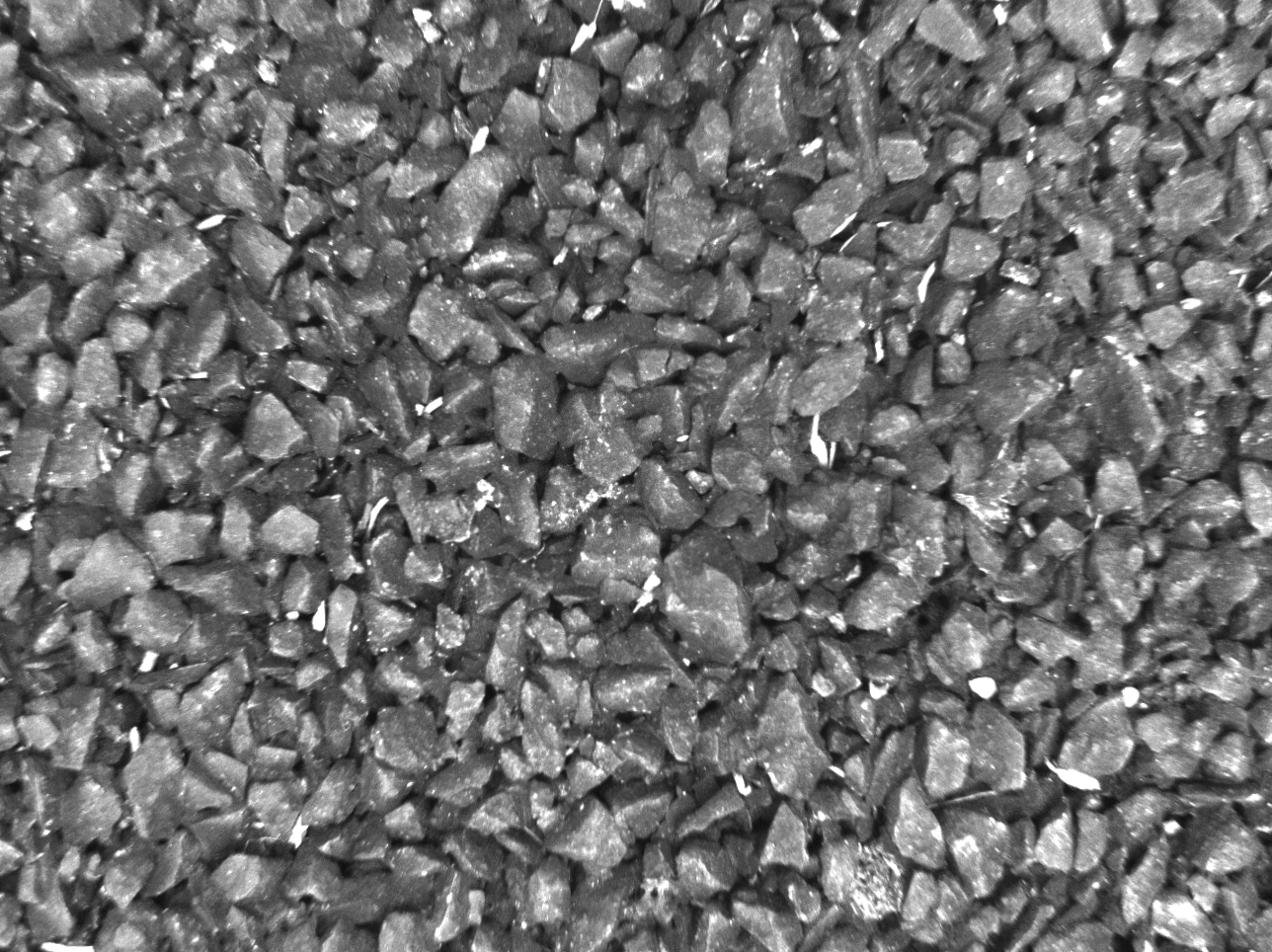}
	\includegraphics[width=0.1614\columnwidth]{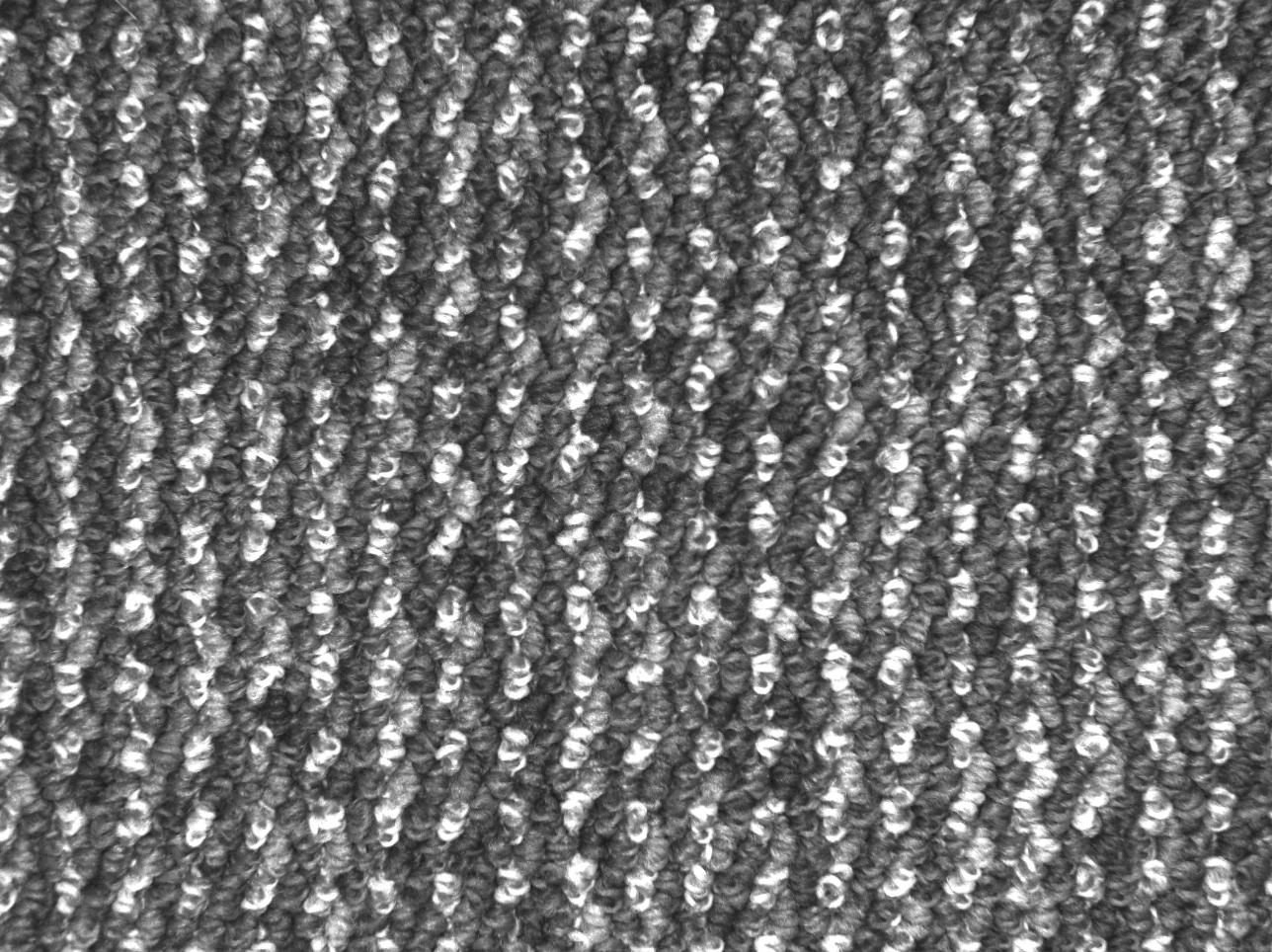}
	\includegraphics[width=0.1614\columnwidth]{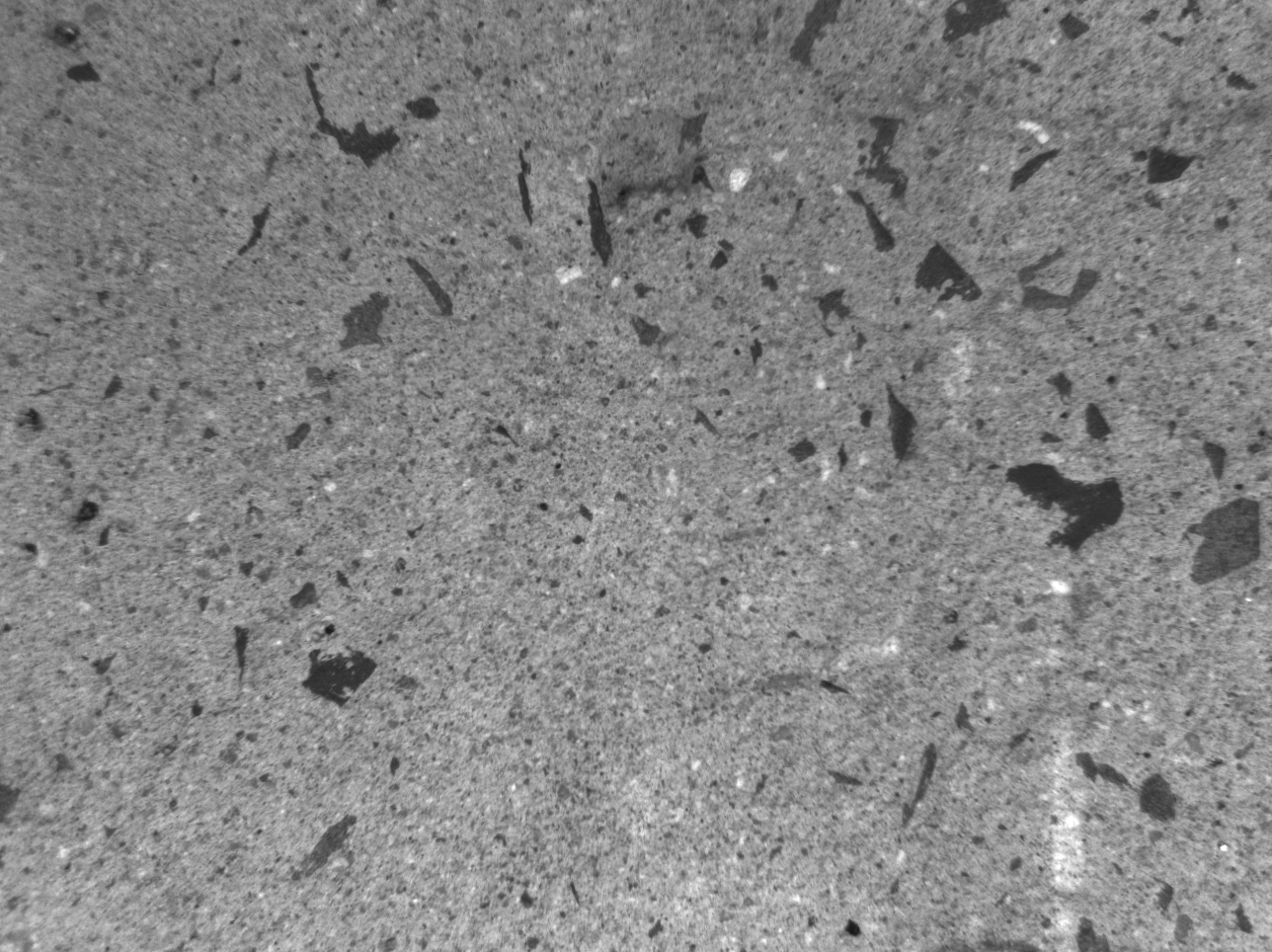}
	\includegraphics[width=0.1614\columnwidth]{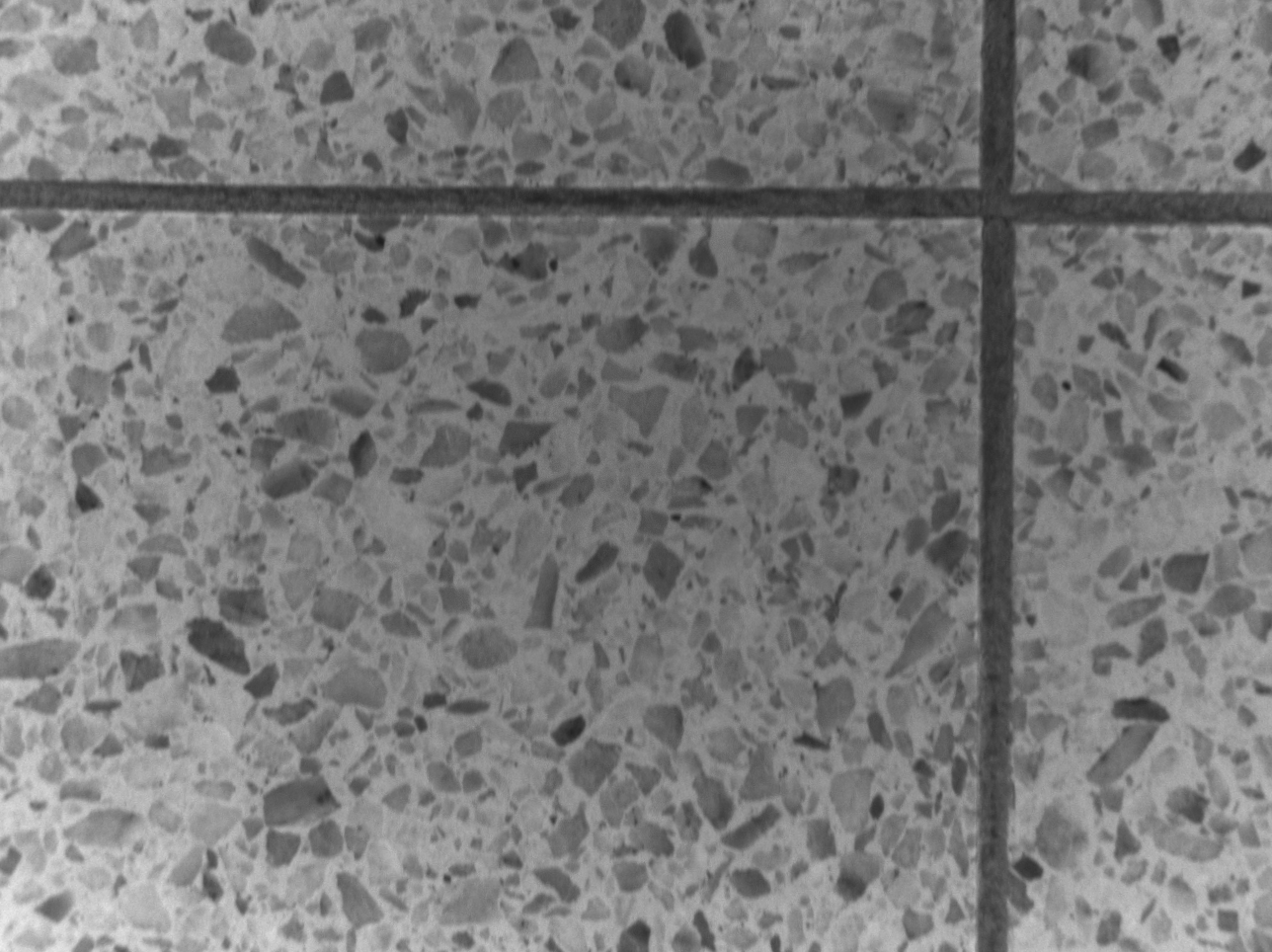}
	\includegraphics[width=0.1614\columnwidth]{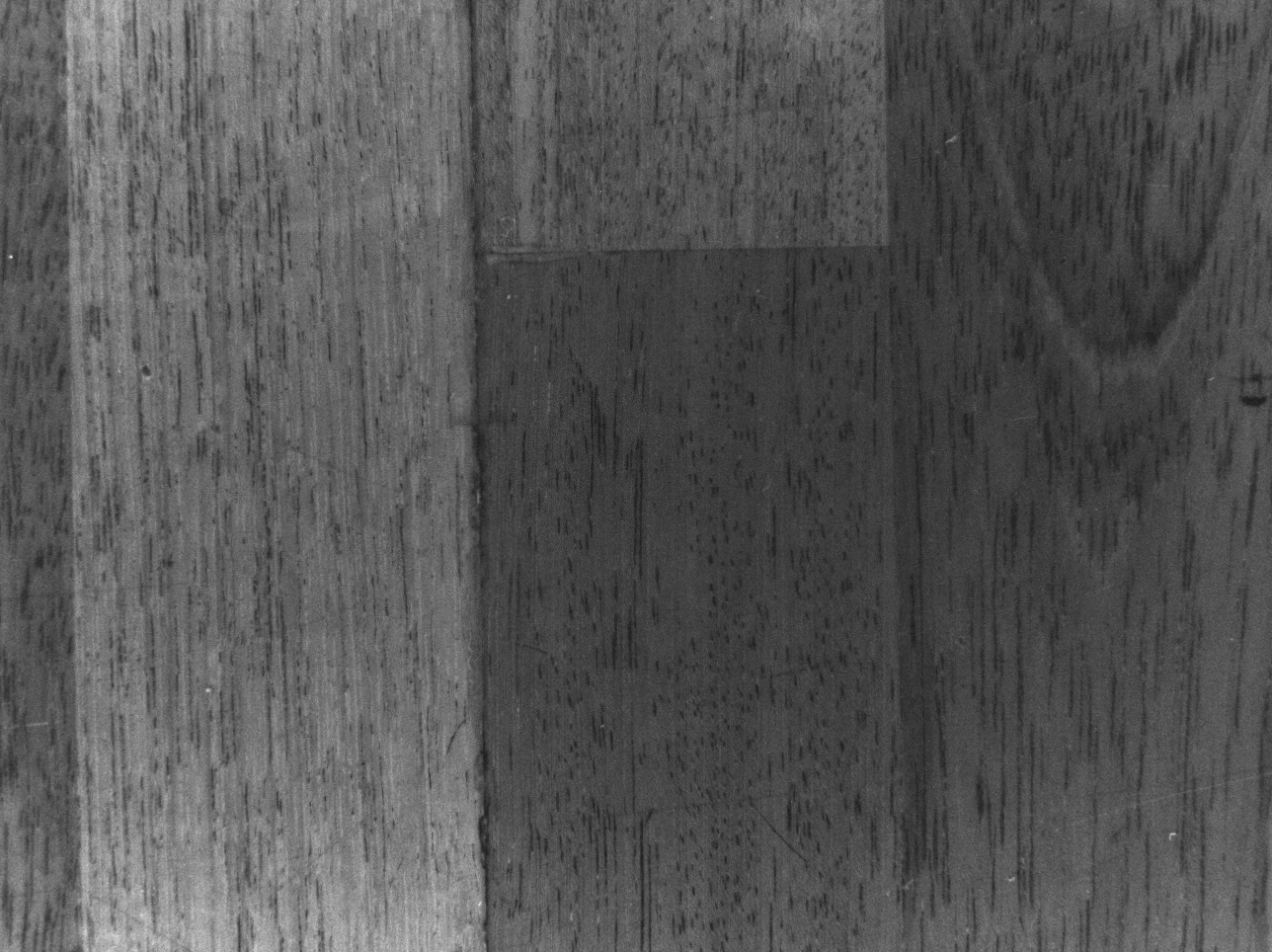}
	\caption{Examples of~\cite{Zhang_High-Prec-Localization}: fine asphalt, coarse asphalt, carpet, concrete, tiles, and wood.}
	\label{fig:example_images}
\end{figure*}

State-of-the-art methods use feature-based localization~\cite{swank2012localization, Nakashima_AKAZE_VO, chenstreetmap, Kozak_Ranger}
that relies on extraction of similar features from varying views of the same location.
While several feature extraction methods were evaluated in these works,
our survey is an extension.
We evaluate additional methods for feature detection (e.g. AKAZE~\cite{Alcantarilla_AKAZE} and LIFT~\cite{Yi_LIFT})
and description (e.g. DAISY~\cite{Tola_DAISY} and LATCH~\cite{Levi_LATCH}),
and we consider different techniques for keypoint selection (non-maximum supression (NMS), adaptive NMS, and bucketing).

This paper contributes an extensive survey using an elaborate evaluation framework for ground texture based localization performance. 
For this purpose, we examine relevant synthetic transformations of ground images,
perform pose estimation in respect to separately taken ground images,
and introduce appropriate performance metrics to evaluate keypoint detector performance on ground images.
Section \ref{sec:localization} presents the localization task based on ground texture and
introduces the evaluated methods.
Then, Section \ref{sec:related_work:feature_evaluation} summarizes other surveys of features for ground texture.
Sections \ref{sec:experimental_setups} and \ref{sec:evaluation} describe and evaluate our experiments. 

\section{Ground texture based localization}
\label{sec:localization}
Ground texture based localization builds on the observation that image patches of ground texture can be used as fingerprint-like identifiers \cite{Zhang_High-Prec-Localization}.
For most application areas, it is reasonable to assume that the ground is flat and therefore that the distance to the ground is known.
Accordingly, with a downward-facing camera,
pose estimation is reduced to determine two coordinates for the position and one orientation angle.
This corresponds to a standard Euclidean transformation of rotation and translation in two dimensions.

Ground texture based localization can be performed with \emph{appearance-based approaches}
\cite{Aqel_Optimal-Config,Zaman_VO,Kelly_AGV, Nagai_Path_Tracking},
e.g. using normalized cross-correlation to find reoccurring image patches,
and with \emph{feature-based approaches} that find feature correspondences
\cite{swank2012localization, Nakashima_AKAZE_VO, Kozak_Ranger, Zhang_High-Prec-Localization,chenstreetmap}.
Furthermore, localization methods can be divided into map-based \emph{absolute localization} and
\emph{incremental localization} for visual odometry \cite{Desouza_VisualLocalizationSurvey}.
Incremental localization can be performed both with
appearance-based \cite{Aqel_Optimal-Config,Zaman_VO} and  with feature-based approaches \cite{swank2012localization, Nakashima_AKAZE_VO}.
If a sufficiently accurate localization prior is available,
appearance-based approaches can be used for absolute localization \cite{Kelly_AGV, Nagai_Path_Tracking}.
However, using image features is potentially more robust to natural degradation of ground texture \cite{Kozak_Ranger}
and can work without localization prior \cite{Zhang_High-Prec-Localization,chenstreetmap}.

\emph{Features} are used to describe characteristic image regions \cite{Pratt_Image-Processing}.
Feature positions are represented by their \emph{keypoints}.
We use the term \emph{keypoint object}, if in addition to the position a size and possibly an orientation is included.
In addition to its keypoint object, features are represented by their \emph{descriptor}, which describes the local environment of a keypoint.

Feature-based localization can be divided into $5$ subtasks:
\begin{enumerate}[noitemsep]
	\item \textbf{Detection:} Find the same keypoint objects from different viewpoints and under varying photometric conditions like illumination, noise, and blur.
	\item \textbf{Selection:} Select a certain number of keypoint objects for further processing.
	\item \textbf{Description:} Compute descriptors that robustly take similar values for corresponding keypoint objects,
	and distinctively different values for non-corresponding ones.
	\item \textbf{Matching:} Propose correspondences between the features found in the current camera image and previously found reference features.
	\item \textbf{Pose estimation:} Based on the proposed correspondences, estimate the current pose.
\end{enumerate}
For the first three tasks, we examine a range of popular approaches available in OpenCV \cite{opencv_library}, as well as LIFT, a state-of-the-art deep learning approach \cite{Yi_LIFT}.
For matching and pose estimation, we revert to standard techniques.
For matching, we compute the Euclidean distance for real-valued descriptors and the Hamming distance for binary ones.
Then, features are matched with linear matching and the \emph{ratio test constraint} as suggested by Lowe~\cite{Lowe_SIFT2}.
This means that for each feature descriptor from the test image the two closest reference descriptors are found.
The closest one is suggested as a match if it is significantly closer than the second one.
Finally, we estimate the relative poses of test images using the proposed feature matches and RANSAC-based estimation of a Euclidean transformation.

\subsection{Evaluated keypoint detectors}
\label{sec:related_work:detection}
Keypoint detection approaches can be split into corner detectors and scale-space detectors~\cite{Agrawal_CenSurE}.
Corners mark suitable keypoints as they tend to be robust to view changes.
The \textbf{Harris} detector~\cite{Harris_Corner} and Good Features To Track (\textbf{GFTT})~\cite{Shi_GFTT}
approximate the second derivative of the sum-of-squared-differences with respect to the shift of a circular image patch to detect edges and corners.
\textbf{FAST}~\cite{Rosten_FAST} compares intensities of center pixels with their surrounding pixels on a circle.
A corner is detected if the circle contains a contiguous sequence of pixels with significantly larger or lower intensity values.
If this condition can no longer be fulfilled it can be rejected early.
To do this, a decision tree defines the order of comparisons.
Mair \etal~\cite{Mair_AGAST} adapt this concept for \textbf{AGAST}.
Instead of using a single decision tree, they switch between multiple ones according to observed local image characteristics.

Scale-space detectors exploit image scale pyramids to find scale invariant keypoints.
Mikolajczyk and Schmid~\cite{Mikolajczyk_Harris_Laplace} extended with \textbf{Harris Laplace} the Harris corner detector
to search for corners in multiple scales using a Gaussian scale-space.
\textbf{SIFT}~\cite{Lowe_SIFT2} detects blobs using a Difference-of-Gaussian (DoG) pyramid as local minima and maxima
of the intensity values in scale and space.
Candidates located on edges or with low contrast are supressed.
Orientation is determined by the dominant local intensity gradients.
\textbf{SURF}~\cite{Bay_SURF} and \textbf{CenSurE}~\cite{Agrawal_CenSurE} approximate the DoG with bi-level Laplacian of Gaussian like Difference-of-Boxes (DoB) or Difference-of-Octagons (DoO), which can be computed efficiently using integral images.
While SIFT and SURF find keypoints using the Hessian measure, CenSurE relies on the Harris corner response.
\textbf{BRISK}~\cite{Leutenegger_BRISK} and \textbf{ORB}~\cite{Rublee_ORB}, on the other hand,
use efficient corner detectors like FAST on a scale pyramid to identify repeatable keypoints in scale-space.
Alcantarilla \etal~\cite{Alcantarilla_KAZE, Alcantarilla_AKAZE} argue that Gaussian scale-space pyramids
and its approximations do not only remove noise, but interesting image details as well.
Therefore, they suggest with \textbf{AKAZE} to find keypoints as maxima of the Hessian in non-linear scale-space.

For \textbf{MSER}~\cite{Matas_MSER} 
the image is thresholded by an increasing illumination value.
Regions with illumination values below the threshold emerge and grow during this process.
Keypoint objects are identified as regions at their point of slowest growth.
In \textbf{MSD}~\cite{Tombari_MSD} image regions that differ from their surrounding in a large neighborhood are considered as keypoint objects.

\subsection{Evaluated keypoint selection methods}
\label{sec:related_work:keypoint_selection}
For feature-based localization it is necessary to extract a certain number of keypoints even on weakly textured images.
Detection parameters should be chosen with respect to this case or need to be adapted texture dependently.
Parameterizing detectors to be able to deal with feature-poor images is difficult.
This problem is emphasized for ground texture images as appearance and frequency of features vary dependent on the type of ground.
Still, using a constant set of parameters is desirable, but results in large numbers of keypoints on feature-rich textures.
Therefore, in order to limit the required processing time for localization,
keypoint selection has an important role for feature extraction on ground texture images.
One approach to keypoint selection is Non-Maximum Suppression (\textbf{NMS}): only the $N$ keypoints with largest response are kept.
In order to improve the spatial distribution of keypoints NMS can be combined with \textbf{bucketing},
where keypoints are detected independently for areas defined by a regular grid~\cite{Kitt_Bucketing}.
An alternative approach is adaptive non-maximum suppression (\textbf{ANMS}),
where keypoints with strong response suppress keypoints in a local neighborhood.

\subsection{Evaluated feature description methods}
\label{sec:related_work:description}
Historically, feature descriptors are real-valued.
\textbf{SIFT}~\cite{Lowe_SIFT2} describes keypoint objects using a histogram of gradient directions.
Similarly, \textbf{DAISY}~\cite{Tola_DAISY} uses quantized orientation histograms.
However, histogram bins are distributed radially around the keypoint and smoothed increasingly with the distance to the keypoint.
\textbf{SURF}~\cite{Bay_SURF} relies on Haar-Wavelet responses that are efficiently to compute using integral images.

More recently, research started to focus on the more compact binary descriptors.
Most of them construct descriptors as concatenated results of pairwise intensity comparisons.
\textbf{BRIEF}~\cite{Calonder_BRIEF1} compares randomly paired pixels from a smoothed image patch.
\textbf{ORB}~\cite{Rublee_ORB} employs a training algorithm to determine a set of pixel comparisons
and rotates this pattern according to the keypoint object orientation.
\textbf{BRISK}~\cite{Leutenegger_BRISK} samples pixel pairs around the keypoint.
While short-distance pairs are evaluated for the descriptor, long-distance pairs are used to determine an orientation.
A similar approach is employed by \textbf{FREAK}~\cite{Alahi_FREAK}, but appropriate intensity comparisons are found in a training process.
In \textbf{LATCH}~\cite{Levi_LATCH} triplets of image patches are compared to each other instead of pixel pairs to increase robustness.
\textbf{AKAZE}~\cite{Alcantarilla_AKAZE} performs pairwise comparisons of first-order gradients.

Most recently, deep learning approaches for feature extraction are developed.
We examine \textbf{LIFT}~\cite{Yi_LIFT}, which is a state-of-the-art method that provides solutions for keypoint detection, orientation estimation, and feature description.
The authors construct an end-to-end trainable Siamese network, which is trained with features from a Structure-from-Motion (SfM) algorithm.
However, in practice they train the network separately for the three tasks.
Training samples consists of four image patches, two corresponding ones for which the network learns to produce similar output and two other patches that should result in distinctively different network outputs.

\section{Surveys of features for ground texture images}
\label{sec:related_work:feature_evaluation}
Zhang \etal~\cite{Zhang_High-Prec-Localization} evaluate the use of SIFT, SURF, ORB, and HardNet~\cite{Mishchuk_HardNet}
for their ground texture based localization pipeline.
The authors receive the best results for keypoint regions and descriptors provided by SIFT.
In a follow-up paper~\cite{Zhang_feature-detector},
the authors develop a fully convolutional neural network trained on ground texture images that achieves higher repeatability than SIFT,
but has increased computational cost.

Kozak \etal~\cite{Kozak_Ranger} evaluate combinations of detector
and descriptor methods on pairs of partially overlapping ground texture images, measuring the number of correctly matched keypoints.
They find the combination of CenSurE keypoints and SIFT descriptors to lead to the largest number of successfully matched features.
Pairings of CenSurE with ORB descriptors, as well as SIFT detector with SIFT descriptor, also show good performance.
FAST, SURF, and GFTT keypoints, as well as descriptors provided by BRISK, FREAK, or SURF,
present significant weaknesses for at least one of the three evaluated road surface texture types: worn asphalt, dark asphalt, and concrete.

Otsu \etal~\cite{Otsu_VO} investigate the suitability of different keypoint detectors for visual odometry from ground texture.
They evaluate Harris, GFTT, and FAST corner detectors as well as the scale-space detectors SIFT, SURF, and CenSurE.
The authors identify that none of the detectors is suited for all situations that occur in the employed desert landscape image datasets.
Therefore, they propose to switch between detectors dependent on the terrain.

This survey extends the prior work.
We evaluate keypoint detection separately like in \cite{Otsu_VO,Zhang_feature-detector},
but also pair them with varying selectors and descriptors.
In addition to the number of correctly matched keypoints \cite{Kozak_Ranger},
we evaluate the repeatability of keypoints and their spatial distribution,
the precision of feature matches, and the pose estimation success rate.
In comparison to \cite{Zhang_High-Prec-Localization}, we evaluate a larger variety of methods for detection and description.
Furthermore, we evaluate performance on synthetically transformed images as well as on separately taken images.
We evaluate sequentially taken image pairs as they occur during incremental localization,
where the transformation is close to a pure translation,
and image pairs taken at different times from independent views as they occur for absolute localization.

\section{Experimental setups}
\label{sec:experimental_setups}
Our experimental framework of three setups is summarized in Table \ref{table:experiments}.
The first experiment examines keypoint detection on synthetic transformations,
the second one feature matching on synthetically transformed images 
and the third one pose estimation using both synthetic transformations and
separately recorded, partially overlapping ground texture image pairs.

\begin{table}
	\begin{center}
		{\scriptsize
			\begin{tabular}{|C{2.67cm}|C{2.45cm}|C{6.4cm}|}%
				\hline
				\bfseries Task & \bfseries Transformation & \bfseries Performance metrics \\\hline
				 Keypoint detection & Synthetic & Repeatability, Ambiguity, < N KPs \\\hline
				 Feature matching & Synthetic & Number of correct matches, Precision\\\hline
				 Pose estimation & Synthetic \& Real & Success rate \\\hline
			\end{tabular}
		}
	\end{center}
	\caption{Experimental setups.}
	\label{table:experiments}
\end{table}

For synthetic transformations,
correct feature matches are known and performance can be evaluated in regard to specific types of image modifications.
We evaluate geometric and photometric transformations.
Typical photometric transformations to consider are Gaussian noise and illumination changes.
The noise is independent and identically distributed (i.i.d.) and zero-mean.
For illumination changes, we employ gamma correction:
pixel values $g$ are modified as: $g_{\mathrm{out}}=\mathrm{round}(g_{\max}\cdot(\frac{g_{\mathrm{in}}}{g_{\max}})^\gamma)$,
where $g_{max}=255$.
Additionally, two geometric transformations are relevant when using downward-facing cameras: rotation and translation.
Rotated images are computed using bicubic interpolation.
In case of translation, an image mask determines a section of the image from which features are extracted.
This mask is translated for testing.
Accordingly, different image sections are evaluated.
For evaluation only keypoints from the intersection between reference mask and test mask are considered.

For separately taken images,
it is difficult to obtain sufficiently accurate ground truth in order to determine which feature matches are correct.
However, it allows us to examine localization performance with its difficulties that occur during application in the real world.
We examine image pairs that are recorded in direct sequence, which represent the challenges of incremental localization,
and we examine image pairs that have been recorded at different times and from independent views,
which represent the challenges of absolute localization.

\subsection{Keypoint detection}
We use synthetic transformations to examine whether the same keypoint objects are found in reference and test image.
Pairs of keypoint objects from reference and test image are considered to match and therefore to represent the same location if
their \emph{Intersection over Union} (IoU) in the reference coordinate frame is greater $0.5$.
As performance metrics, we evaluate keypoint \textbf{repeatability} introduced by Mikolajczyk~\etal~\cite{Mikolajczyk_Repeatability}.
It measures the proportion of keypoints from the test image that were also found in the reference image.
Additionally, we introduce two novel performance metrics.
\textbf{Ambiguity} addresses the problem of repeatability that it does not penalize ambiguous keypoint correspondences.
This problem occurs if a keypoint object from the test image has multiple matches in the reference image,
which happens if keypoints are clustered.
We compute ambiguity as the mean number of matches of test image keypoint objects with at least one match.
Therefore, $\mathrm{ambiguity} \ge 1.0$.
An ambiguity greater $1.0$ suggests that the repeatability score is inflated by ambiguous keypoint matches.
The second new metric, \textbf{< N KPs}, shows how often fewer than $N$ keypoints are found,
as having only few keypoints increases the risk of failure for feature-based localization.

\subsection{Feature matching}
In order to evaluate whether the obtained features are suited for the localization pipeline, we examine feature matching performance.
We evaluate the \textbf{number of correctly matched features} and compute the the matching \textbf{precision}:
$\mathrm{precision} = \frac{\mathrm{\#correct~matches}}{\mathrm{\#correct~matches} + \mathrm{\#incorrect~matches}}$.

\subsection{Pose estimation}
Pose estimates are considered correct if
their distance to the \textit{ground truth} is less than $30$ pixels ($4.8$\,mm) and 
if their orientation error is less than $\pm1.5$ degrees (these thresholds are adopted from Zhang~\etal~\cite{Zhang_High-Prec-Localization}).
We evaluate pose estimation performance using the \textbf{success rate} metric,
which is computed as the ratio of correct pose estimates to incorrect ones.

\section{Evaluation}
\label{sec:evaluation}
For evaluation of our experimental framework,
we use the ground texture image database of Zhang~\etal~\cite{Zhang_High-Prec-Localization}.
We test on all six ground texture types captured by a gray-scale camera (see Figure \ref{fig:example_images}).
Images have a size of $1288$ by $964$ pixels.
We select $3$ images per texture to be used exclusively for parameter optimization, $100$ for synthetic transformations, 
and $100$ image pairs each for incremental and absolute localization tasks.
We observed no significant performance variations using more test images.
Our strategies for parameter optimization,
and the obtained parameter settings can be found in the \emph{supplementary material}.

We make use of OpenCV $4.0$~\cite{opencv_library} implementations for most of the evaluated detectors and descriptors.
Due to bad performance of the ORB implementation of OpenCV, we use its implementation that comes with ORB-SLAM2 \cite{Mur-Artal_ORBSLAM2}.
The implementation and the trained network weights of LIFT are provided by the authors, 
which claim to achieve good generalization performance even without domain specific training samples \cite{Yi_LIFT}.
We exclude ORB and LIFT from the evaluation on synthetic transformations as they do not allow to define a detection mask.
For feature matching, we find most similar reference descriptors and filter them with the ratio test constraint with a threshold of $0.7$.
Poses are estimated using RANSAC-based estimation of a Euclidean transformation with $2000$ iterations and the error threshold applied in \cite{Zhang_High-Prec-Localization} of $3.0$.
We observed better localization performance estimating not only the three obligatory parameters for position and orientation,
but also an additional scale parameter allowing for small variations in height.

Synthetic transformations are parametrized as follows:
for rotation, angles between $0$ and $180$ degrees; for translation,
the detection mask of the test image is pushed in direction of the lower right image corner (in respect to the reference mask),
the resulting IoUs of reference mask and test mask are between $0.2$ and $1.0$.
Gaussian i.i.d. noise is zero-mean with standard deviation between $0.0$ and $40.0$;
illumination values are changed using a gamma between $0.1$ and $3.0$.
When presenting results from synthetic transformations,
metrics are averaged with equal contribution of the results of each transformation type.
Transformation and texture dependent results are presented in the \emph{supplementary material}.

\subsection{Evaluation of selector-detector pairings}
We examine the repeatability of keypoint detectors using the keypoint selection methods introduced in Section \ref{sec:related_work:keypoint_selection} to reduce the number of keypoints to $1000$.
Respectively, if the keypoint detection method allows to specify the desired number of keypoints,
we set this parameter to $1000$.
For ANMS, we use Suppression via Square Covering (SSC) \cite{Bailo_SSC} with a tolerance of $20\%$.
For bucketing, we received good results for non-square buckets, using a grid of $8$ rows and $6$ columns.
For each grid cell $21$ keypoints are selected using NMS resulting in a maximum of $1008$ keypoints per image.
Evaluating a reference image without selection with all our synthetic transformations takes us several days,
which is why for this experiment we evaluate a single test image per type of ground texture.
In addition to the repeatability scores,
Table \ref{table:keypoint_selector_results_synth} presents the average number of keypoints before selection.
MSER does not provide a keypoint response measure, and is therefore not well suited to be used with a selection method.
Together with FAST, AGAST, and BRISK, MSER has significantly better repeatability without selection.
In order to select MSER keypoints, we use the order of extracted keypoints as substitution for the response measure.
This means that the first found maximally stable extremal regions, which are the ones with lowest intensity values,
are considered to have the largest response.
With this work-around MSER still achieves surprisingly large repeatability of $51\%$ using NMS and $73\%$ using ANMS.
We find MSER to be the only detector that performs best with ANMS.
For all other detectors, we use NMS in the following.
The repeatability of SURF and especially MSD is increased when using keypoints selected by NMS instead of all available keypoints.
This means that keypoints that have been assigned low values of the response measures of these detectors
are indeed non-repeatable and are rightly removed by NMS.

\begin{table}
	\begin{center}
		{\scriptsize 
			\begin{tabular}{|c|c|c|c|c|c|}
				\hline
				Detector & $\#$KPs before selection & Repeatability w/o selection & Rep. NMS & Rep. ANMS & Rep. Bucketing\\\hline\hline
				AKAZE & 10199 & 0.81 & \textbf{0.82} & 0.41 & 0.74 \\\hline
				SIFT & 755 & 0.82 & \textbf{0.82} & \textbf{0.82} & 0.77 \\\hline
				SURF & 7271 & 0.80 & \textbf{0.82} & 0.65 & 0.74 \\\hline
				CenSurE & 6434 & 0.83 & \textbf{0.76} & 0.39 & 0.70 \\\hline				
				MSD & 6484 & 0.59 & \textbf{0.76} & 0.51 & 0.68 \\\hline
				HarrisLaplace & 839 & 0.76 & \textbf{0.76} & \textbf{0.76} & 0.68 \\\hline
				MSER & 15238 & 0.94 & 0.51 & \textbf{0.73} & 0.52 \\\hline
				BRISK & 58050 & 0.84 & \textbf{0.71} & 0.25 & 0.64 \\\hline
				GFTT & 894 & 0.69 & \textbf{0.69} & \textbf{0.69} & 0.29 \\\hline
				AGAST & 225361 & 0.93 & \textbf{0.64} & 0.22 & 0.57 \\\hline
				FAST & 52112 & 0.78 & \textbf{0.64} & 0.26 & 0.56 \\\hline				
			\end{tabular}
		}
	\end{center}
	\caption{Number of keypoints before selection and repeatability of selector-detector pairings.}
	\label{table:keypoint_selector_results_synth} 
\end{table}

In a next step, we evaluate the best performing detector-selector pairings using all $100$ test reference images.
Results are presented in Table \ref{table:keypoint_detector_results}.
For the < N KPs metric, we set $N$ to $100$, as we noticed that localization success is low with fewer keypoints.
For most detectors, we were able to find parameters that allow to retrieve at least $100$ keypoints from almost all images.
But, AKAZE, Harris Laplace, and GFTT still find fewer keypoints on at least $4\%$ of the images.
This problem occurs almost exclusively on wood texture.
Harris Laplace extracts less than $100$ keypoints on $49\%$ of wood images, GFTT on $28\%$ and AKAZE on $23\%$.
SIFT, AKAZE, and SURF have with $83\%$ to $84\%$ the best repeatability.
However, SIFT has a large ambiguity score of $1.5$,
as it detects multiple orientations for some keypoints.

Overall, our evaluation suggests that SURF and CenSurE, as well as AKAZE for non-wood texture,
are the best detectors on ground texture images.
They have among the best repeatability, and ambiguity scores,
and are, unlike SIFT and MSD, fast to compute.

\begin{table}
	\begin{center}
		{\scriptsize
			\csvreader[tabular=|c|c|c|c|c|c|c|,
			table head=\hline Selector & Detector & < 100 KPs & Repeatability & Ambiguity & Computation time (s)\\\hline\hline,
			late after line=\\\hline]%
			{csv/keypoint_detector_results.csv}{Detector=\Detector, keypointSelector=\keypointSelector, LessThanHundret=\LessThanHundret, AdjustedRepeatability=\AdjustedRepeatability, Repeatability=\Repeatability, Coverage=\Coverage, ResponseRatio=\ResponseRatio, MeanNOFoverlappingKeypoints=\MeanNOFoverlappingKeypoints, DetectionTime=\DetectionTime}%
			{\keypointSelector & \Detector & \LessThanHundret & \Repeatability & \MeanNOFoverlappingKeypoints & \DetectionTime}%
		}
	\end{center}
	\caption{Evaluation of keypoint detectors on synthetically transformed images.}
	\label{table:keypoint_detector_results} 
\end{table}

\subsection{Evaluation of detector-descriptor pairings}
We evaluate pose estimation success rate for all working detector-descriptor pairs.
The AKAZE descriptor only allows to use AKAZE keypoints.
DAISY requires keypoint objects to specify orientation.
The ORB descriptor has requirements on the keypoint scaling,
which excludes SIFT and LIFT.
We evaluate on image pairs from incremental localization (Table \ref{table:detector_descriptor_results_seq}), 
as well as on the image pairs from absolute localization (Table \ref{table:detector_descriptor_results_abs}).
The intersection of the sequentially taken image pairs is on average $22.7\%$.
Some almost non-overlapping pairs with intersections as low as $1.7\%$ are particularly challenging.
The image pairs from the absolute localization tasks have larger intersections with an average of $43.7\%$.
However, in this case the rotation between the images is with an average of $120$ degrees higher as for the pairs from incremental localization with an average rotation of $3$ degrees.
Again, the number of retrieved keypoints is reduced to $1000$ using the best selection method.
The best performance for incremental localization of $93\%$ success rate is achieved with ORB on CenSurE or MSD keypoints.
BRIEF and LATCH perform similarly well with $90\%$ success rate.

For absolute localization most descriptors can achieve more than $90\%$ success rate if paired with the right detector.
SURF and DAISY are not quite as successful as they struggle with images of wooden texture.
Detectors that provide orientation information (SIFT, SURF, AKAZE, ORB, BRISK, and LIFT)
outperform the other detectors.
ORB, BRIEF, LATCH, SIFT, and LIFT descriptors rely on available orientation information and perform poorly if it is missing.
In these cases pose estimation success rate drops to about $10\%$ to $15\%$.

\begin{table}
	\begin{center}
		{\scriptsize
			\setlength{\tabcolsep}{5.8pt} 
			\begin{tabular}{|c||c|c|c|c|c|c|c|c|c|c|}
				\hline
				\multirow{2}{*}{Detector} & \multicolumn{10}{c|}{Descriptor}\\\cline{2-11}
				& ORB & BRIEF & LATCH & SURF & SIFT & AKAZE & LIFT & BRISK & FREAK & DAISY \\\hline\hline
				CenSurE & \textbf{0.93} & \textbf{0.90} & \textbf{0.90} & 0.24 & 0.82 & NA & \textbf{0.84} & 0.80 & 0.70 & NA \\\hline
				MSD & \textbf{0.93} & \textbf{0.90} & \textbf{0.90} & 0.68 & 0.83 & NA & \textbf{0.84} & \textbf{0.81} & 0.69 & NA \\\hline
				HarrisLaplace & 0.88 & 0.86 & 0.82 & 0.75 & 0.60 & NA & 0.62 & 0.74 & 0.67 & NA \\\hline
				SURF & 0.89 & 0.86 & 0.86 & \textbf{0.87} & 0.73 & NA & 0.40 & 0.72 & 0.73 & \textbf{0.72} \\\hline
				AKAZE & 0.80 & 0.85 & 0.83 & 0.47 & 0.73 & \textbf{0.84} & 0.76 & 0.80 & \textbf{0.74} & 0.66 \\\hline
				FAST & 0.89 & 0.85 & 0.85 & 0.24 & 0.79 & NA & 0.77 & 0.76 & 0.66 & NA \\\hline
				GFTT & 0.88 & 0.84 & 0.84 & 0.21 & 0.80 & NA & 0.80 & 0.79 & 0.68 & NA \\\hline
				LIFT & NA & 0.84 & 0.83 & 0.36 & 0.69 & NA & 0.75 & 0.80 & 0.72 & 0.66 \\\hline
				MSER & 0.83 & 0.83 & 0.85 & 0.59 & 0.76 & NA & 0.47 & 0.62 & 0.56 & NA \\\hline
				SIFT & NA & 0.82 & 0.84 & 0.59 & \textbf{0.84} & NA & 0.61 & 0.70 & 0.63 & 0.69 \\\hline
				AGAST & 0.77 & 0.76 & 0.75 & 0.30 & 0.66 & NA & 0.61 & 0.64 & 0.55 & NA \\\hline
				ORB & 0.70 & 0.71 & 0.71 & 0.73 & 0.58 & NA & 0.12 & 0.49 & 0.62 & 0.47 \\\hline
				BRISK & 0.66 & 0.69 & 0.70 & 0.61 & 0.61 & NA & 0.38 & 0.71 & 0.66 & 0.48 \\\hline
			\end{tabular}
		}
	\end{center}
	\caption{Pose estimation success rate on image pairs from incremental localization tasks, where images are taken in direct sequence.}
	\label{table:detector_descriptor_results_seq} 
\end{table}

\begin{table}
	\begin{center}
		{\scriptsize 
			\setlength{\tabcolsep}{5.8pt} 
			\begin{tabular}{|c|c|c|c|c|c|c|c|c|c|c|}
				\hline
				\multirow{2}{*}{Detector} & \multicolumn{10}{c|}{Descriptor}\\\cline{2-11}
				& ORB & BRIEF & LATCH & SURF & SIFT & AKAZE & LIFT & BRISK & FREAK & DAISY \\\hline\hline
				CenSurE & 0.14 & 0.11 & 0.09 & 0.50 & 0.10 & NA & 0.11 & 0.88 & 0.96 & NA \\\hline
				MSD & 0.14 & 0.11 & 0.09 & 0.72 & 0.09 & NA & 0.10 & 0.86 & 0.90 & NA \\\hline
				HarrisLaplace & 0.14 & 0.10 & 0.09 & 0.71 & 0.10 & NA & 0.10 & 0.87 & 0.89 & NA \\\hline
				SURF & 0.89 & 0.98 & 0.98 & \textbf{0.87} & 0.94 & NA & 0.91 & \textbf{0.96} & \textbf{0.98} & 0.70 \\\hline
				AKAZE & \textbf{0.94} & \textbf{0.99} & \textbf{0.99} & 0.73 & 0.96 & \textbf{0.99} & \textbf{0.96} & 0.92 & \textbf{0.98} & \textbf{0.75} \\\hline
				FAST & 0.14 & 0.11 & 0.09 & 0.46 & 0.10 & NA & 0.11 & 0.84 & 0.89 & NA \\\hline
				GFTT & 0.14 & 0.11 & 0.09 & 0.37 & 0.10 & NA & 0.11 & 0.86 & 0.89 & NA \\\hline				
				LIFT & NA & 0.71 & 0.85 & 0.17 & 0.70 & NA & 0.52 & 0.46 & 0.73 & 0.24 \\\hline
				MSER & 0.13 & 0.10 & 0.09 & 0.76 & 0.09 & NA & 0.10 & 0.90 & 0.89 & NA \\\hline				
				SIFT & NA & 0.97 & \textbf{0.99} & 0.75 & \textbf{0.99} & NA & \textbf{0.96} & 0.93 & 0.94 & 0.74 \\\hline
				AGAST & 0.12 & 0.10 & 0.09 & 0.46 & 0.08 & NA & 0.10 & 0.80 & 0.86 & NA \\\hline
				ORB & 0.82 & 0.86 & 0.89 & 0.84 & 0.90 & NA & 0.73 & 0.84 & 0.87 & 0.56 \\\hline
				BRISK & 0.79 & 0.88 & 0.89 & 0.76 & 0.84 & NA & 0.84 & 0.87 & 0.89 & 0.53 \\\hline
			\end{tabular}
		}
	\end{center}
	\caption{Pose estimation success rate on image pairs from absolute localization tasks, where images are taken at different times and from varying perspectives.}
	\label{table:detector_descriptor_results_abs}
\end{table}

For further analysis of feature descriptor performance,
we use synthetic transformations to evaluate the pairings of detectors and descriptors that performed the best on absolute localization.
In cases of multiple best performing detectors, we use the faster one.
Table \ref{table:keypoint_descriptor_results_synth} presents results for feature matching and pose estimation.
Additionally, we provide the results of BRIEF on CenSurE keypoints.
We note that the challenges of our synthetic transformations, which include severe rotations and photometric modifications,
are more similar to the ones of absolute localization.
Accordingly, BRIEF has significantly better performance using AKAZE keypoint objects instead of CenSurE keypoint objects, which lack orientation information.
SIFT outperforms the other feature extraction pipelines.
We find precision to correlate with the pose estimation success rate,
while this relation is not that clear for the number of correct matches.
E.g. BRIEF on CenSurE has about $50$ more correct matches than DAISY on AKAZE,
despite having a significantly lower pose estimation success rate.

\begin{table}
	\begin{center}
		{\scriptsize 
			\begin{tabular}{|c|c|c|c|c|}
				\hline
				Detector &Descriptor & Pose estimation success rate & \#Correct matches & Precision \\\hline\hline
				SIFT & SIFT & 1.00 & 551 & 1.00 \\\hline
				SURF & BRISK & 0.99 & 559 & 1.00 \\\hline
				AKAZE & AKAZE & 0.98 & 545 & 0.99 \\\hline
				AKAZE & BRIEF & 0.98 & 534 & 0.99 \\\hline
				AKAZE & FREAK & 0.98 & 561 & 0.99 \\\hline
				AKAZE & LATCH & 0.97 & 538 & 0.98 \\\hline
				SURF & SURF & 0.97 & 509 & 0.99 \\\hline
				AKAZE & DAISY & 0.87 & 410 & 0.98 \\\hline
				CenSurE & BRIEF & 0.77 & 460 & 0.92 \\\hline
			\end{tabular}
		}
	\end{center}
	\caption{Evaluation of detector-descriptor pairings on synthetically transformed images.}
	\label{table:keypoint_descriptor_results_synth}
\end{table}

Table \ref{table:success_rate_against_ref_KPs} presents success rates for consecutive image pairs using different numbers of reference image features.
Our selection of $N=100$ for the < 100 KPs metric is validated,
as localization performance is low for $100$ or less reference keypoints.
On the other hand, pairings like CenSurE-ORB, CenSurE-LATCH, and SIFT-SIFT reach values close to their best performance at $300$ features already.
Others, like SURF-SURF and SURF-DAISY should not be used with less than $500$ reference features.
Furthermore, we find further evidence for our observation that the number of correct matches is not a suitable indicator for localization performance.
While the number of RANSAC inliers (and therefore the number of correct matches) increases with the number of reference features,
the localization success rates stagnate at some point.
For CenSurE-BRIEF, for example, the number of inliers in successful localization attempts increases from about $35$ at $300$ reference features to $94$ for $1500$ reference features, even though success rate increases only from $87\%$ to $90\%$.
Once a certain number of correct matches is available, localization performance does not increase further.

\begin{table}
	\begin{center}
		{\scriptsize 
			\begin{tabular}{|c|c|c|c|c|c|c|c|c|c|c|}
				\hline
				Detector & Descriptor & 10 & 50 & 100 & 200 & 300 & 500 & 750 & 1000 & 1500 \\\hline\hline
					CenSurE & ORB & 0.01 & 0.54 & 0.75 & 0.87 & 0.89 & 0.91 & 0.92 & 0.92 & 0.92 \\\hline
					CenSurE & BRIEF & 0.00 & 0.47 & 0.73 & 0.84 & 0.87 & 0.88 & 0.89 & 0.89 & 0.90 \\\hline
					CenSurE & LATCH & 0.13 & 0.65 & 0.77 & 0.84 & 0.86 & 0.88 & 0.88 & 0.89 & 0.89 \\\hline
					SURF & SURF & 0.01 & 0.25 & 0.36 & 0.56 & 0.69 & 0.83 & 0.84 & 0.86 & 0.86 \\\hline
					SIFT & SIFT & 0.12 & 0.56 & 0.71 & 0.80 & 0.82 & 0.83 & 0.84 & 0.83 & 0.84 \\\hline
					CenSurE & LIFT & 0.04 & 0.50 & 0.67 & 0.76 & 0.80 & 0.82 & 0.82 & 0.83 & 0.83 \\\hline
					AKAZE & AKAZE & 0.03 & 0.41 & 0.62 & 0.73 & 0.77 & 0.81 & 0.83 & 0.83 & 0.83 \\\hline
					MSD & BRISK & 0.03 & 0.28 & 0.46 & 0.69 & 0.74 & 0.79 & 0.81 & 0.80 & 0.80 \\\hline
					AKAZE & FREAK & 0.01 & 0.33 & 0.46 & 0.59 & 0.65 & 0.71 & 0.72 & 0.73 & 0.74 \\\hline
					SURF & DAISY & 0.00 & 0.16 & 0.19 & 0.31 & 0.45 & 0.65 & 0.70 & 0.71 & 0.71 \\\hline
			\end{tabular}
		}
	\end{center}
	\caption{Pose estimation success rates for varying numbers of reference features.}
	\label{table:success_rate_against_ref_KPs}
\end{table}

\section{Conclusions}
\label{sec:conclusions}
We examined keypoint detectors, selection methods, and feature descriptors on synthetically transformed ground images as well as on pairs of separately taken ground images.

In contrast to Otsu~\etal~\cite{Otsu_VO}, we find with SURF and CenSurE keypoint detectors that are well suited for all evaluated ground textures.
For image pairs where the transformation consists mainly of a translation, as it is the case for the task of incremental localization,
we can confirm the suitability of ORB descriptors on CenSurE keypoints as well as SIFT features,
and the weaknesses of FAST and GFTT keypoints, and BRISK and FREAK descriptors,
as assessed by Kozak and Alban \cite{Kozak_Ranger}.
This is even though our evaluation has shown that their metric, the number of correctly matched features,
is not necessarily a good indicator for localization performance.
SURF, on the other hand, has shown good performance for us.
Finally, we validated the observation of Zhang~\etal~\cite{Zhang_High-Prec-Localization} that SIFT is suited for absolute localization as it is among the best performing methods for the estimation of transformations between image pairs that have been taken at different times and perspectives,
and the best feature extractor to deal with even more severe synthetic transformations.
However, other pairings like BRIEF, LATCH, and AKAZE descriptors on AKAZE keypoints perform similarly well and are significantly faster to compute.
For further research, we are interested in absolute localization performance using different matching and pose estimation methods.

\bibliography{./bib}

\end{document}